\documentclass{article}

\usepackage{microtype}
\usepackage{graphicx}
\usepackage{subfigure}
\usepackage{booktabs} 

\usepackage{hyperref}

\usepackage{algpseudocode}
\usepackage{amssymb, amsmath}

\usepackage{amsmath,amsfonts,bm}

\def\eqref#1{equation~\ref{#1}}

\def\1{\bm{1}}

\def\eps{{\epsilon}}

\DeclareMathAlphabet{\mathsfit}{\encodingdefault}{\sfdefault}{m}{sl}
\SetMathAlphabet{\mathsfit}{bold}{\encodingdefault}{\sfdefault}{bx}{n}

\usepackage{siunitx}

\usepackage[accepted]{icml2021}

\icmltitlerunning{Decoupling Representation Learning from Reinforcement Learning}

\begin{document}

\twocolumn[
\icmltitle{Decoupling Representation Learning from 
           Reinforcement Learning}



\icmlsetsymbol{equal}{*}

\begin{icmlauthorlist}
\icmlauthor{Adam Stooke}{berk}
\icmlauthor{Kimin Lee}{berk}
\icmlauthor{Pieter Abbeel}{berk}
\icmlauthor{Michael Laskin}{berk}
\end{icmlauthorlist}

\icmlaffiliation{berk}{University of California, Berkeley}

\icmlcorrespondingauthor{Adam Stooke}{adam.stooke@berkeley.edu}
\icmlcorrespondingauthor{Michael Laskin}{mlaskin@berkeley.edu}

\icmlkeywords{Machine Learning, ICML}

\vskip 0.3in
]



\printAffiliationsAndNotice{}  

\begin{abstract}
In an effort to overcome limitations of reward-driven feature learning in deep reinforcement learning (RL) from images, we propose decoupling representation learning from policy learning.  To this end, we introduce a new unsupervised learning (UL) task, called Augmented Temporal Contrast (ATC), which trains a convolutional encoder to associate pairs of observations separated by a short time difference, under image augmentations and using a contrastive loss.  In online RL experiments, we show that training the encoder exclusively using ATC matches or outperforms end-to-end RL in most environments.  Additionally, we benchmark several leading UL algorithms by pre-training encoders on expert demonstrations and using them, with weights frozen, in RL agents; we find that agents using ATC-trained encoders outperform all others.  We also train multi-task encoders on data from multiple environments and show generalization to different downstream RL tasks.  Finally, we ablate components of ATC, and introduce a new data augmentation to enable replay of (compressed) latent images from pre-trained encoders when RL requires augmentation.  Our experiments span visually diverse RL benchmarks in DeepMind Control, DeepMind Lab, and Atari, and our complete code is available at \url{https://github.com/astooke/rlpyt/tree/master/rlpyt/ul}.
\end{abstract}

\section{Introduction}
Ever since the first fully-learned approach succeeded at playing Atari games from screen images \citep{mnih2015human}, standard practice in deep reinforcement learning (RL) has been to learn visual features and a control policy jointly, end-to-end.  Several such deep RL algorithms have matured \citep{hessel2017rainbow,schulman2017proximal,a3c,haarnoja2018soft} and have been successfully applied to domains ranging from real-world \citep{levine2015end,kalashnikov2018qt} and simulated robotics \citep{lee2019stochastic,laskin_lee2020rad,hafner2019dream} to sophisticated video games \citep{openai2019dota,jaderberg2019human}, and even high-fidelity driving simulators \citep{dosovitskiy2017}.  While the simplicity of end-to-end methods is appealing, relying on the reward function to learn visual features can be severely limiting.  For example, it leaves features difficult to acquire under sparse rewards, and it can narrow their utility to a single task.  Although our intent is broader than to focus on either sparse-reward or multi-task settings, they arise naturally in our studies.  We investigate how to learn visual representations which are agnostic to rewards, without degrading the control policy.

A number of recent works have significantly improved RL performance by introducing auxiliary losses, which are unsupervised tasks that provide feature-learning signal to the convolution neural network (CNN) encoder, additionally to the RL loss \citep{jaderberg2016reinforcement,oord2018representation,laskin2020curl,guo2020bootstrap,schwarzer2020}.  Meanwhile, in the field of computer vision, recent efforts in unsupervised and self-supervised learning \citep{chen2020simclr,grill2020bootstrap,kaiming2019moco} have demonstrated that powerful feature extractors can be learned without labels, as evidenced by their usefulness for downstream tasks such as ImageNet classification. Together, these advances suggest that visual features for RL could possibly be learned entirely without rewards, which would grant greater flexibility to improve overall learning performance.  To our knowledge, however, no single unsupervised learning (UL) task has been shown adequate for this purpose in general vision-based environments.

In this paper, we demonstrate the first decoupling of representation learning from reinforcement learning that performs as well as or better than end-to-end RL.  We update the encoder weights using only UL and train a control policy independently, on the (compressed) latent images.  This capability stands in contrast to previous state-of-the-art methods, which have trained the UL and RL objectives jointly, or \cite{laskin2020curl}, which observed diminished performance with decoupled encoders.

Our main enabling contribution is a new unsupervised task tailored to reinforcement learning, which we call Augmented Temporal Contrast (ATC).  ATC requires a model to associate observations from nearby time steps within the same trajectory \citep{anand2019unsupervised}.  Observations  are encoded via a convolutional neural network (shared with the RL agent) into a small latent space, where the InfoNCE loss is applied \citep{oord2018representation}.  Within each randomly sampled training batch, the positive observation, $o_{t+k}$, for every anchor, $o_t$, serves as negative for all other anchors.  For regularization, observations undergo stochastic data augmentation \citep{laskin2020curl} prior to encoding, namely random shift \citep{kostrikov2020image}, and a momentum encoder \citep{he2019momentum,laskin2020curl} is used to process the positives.  A learned predictor layer further processes the anchor code \citep{grill2020bootstrap,chen2020simclr} prior to contrasting.  In summary, our algorithm is a novel combination of elements that enables generic learning of the structure of observations and transitions in MDPs without requiring rewards or actions as input. 


We include extensive experimental studies establishing the effectiveness of our algorithm in a visually diverse range of common RL environments: DeepMind Control Suite (DMControl; \citealt{tassa2018deepmind}), DeepMind Lab (DMLab; \citealt{dmlab}), and Atari \citep{bellemare2013arcade}.  Our experiments span discrete and continuous control, 2D and 3D visuals, and both on-policy and off policy RL algorithms.  Complete code for all of our experiments is available at \url{hiddenurl}.  Our empirical contributions are summarized as follows:


\textit{Online RL with UL}:  We find that the convolutional encoder trained solely with the unsupervised ATC objective can fully replace the end-to-end RL encoder without degrading policy performance. ATC achieves nearly equal or greater performance in all DMControl and DMLab environments tested and in 5 of the 8 Atari games tested. In the other 3 Atari games, 
using ATC as an auxiliary loss or for weight initialization still brings improvements over end-to-end RL. 

\textit{Encoder Pre-Training Benchmarks}:  We pre-train the convolutional encoder to convergence on expert demonstrations, and evaluate it by training an RL agent using the encoder with weights frozen.  We find that ATC matches or outperforms all prior UL algorithms as tested across all domains, demonstrating that ATC is a state-of-the-art UL algorithm for RL.

\textit{Multi-Task Encoders}: An encoder is trained on demonstrations from multiple environments, and is evaluated, with weights frozen, in separate downstream RL agents.  A single encoder trained on four DMControl environments generalizes successfully, performing equal or better than end-to-end RL in four held-out environments.  Similar attempts to generalize across eight diverse Atari games result in mixed performance, confirming some limited feature sharing among games.

\textit{Ablations and Encoder Analysis}: Components of ATC are ablated, showing their individual effects.  Additionally, data augmentation is shown to be necessary in DMControl during RL even when using a frozen encoder.  We introduce a new augmentation, \emph{subpixel random shift}, which matches performance while augmenting the latent images, unlocking computation and memory benefits.

\section{Related Work}

Several recent works have used unsupervised/self-supervised representation learning methods to improve performance in RL.  The UNREAL agent \citep{jaderberg2016reinforcement} introduced unsupervised auxiliary tasks to deep RL, including the Pixel Control task, a Q-learning method requiring predictions of screen changes in discrete control environments, which has become a standard in DMLab \citep{hessel2019multi}.  CPC \citep{oord2018representation} applied contrastive losses over multiple time steps as an auxiliary task for the convolutional and recurrent layers of RL agents, and it has been extended with future action-conditioning \citep{guo2018neural}.  Recently, PBL \citep{guo2020bootstrap} surpassed these methods with an auxiliary loss of forward and backward predictions in the recurrent latent space using partial agent histories. Where the trend is of increasing sophistication in auxiliary recurrent architectures, our algorithm is markedly simpler, requiring only observations, and yet it proves sufficient in partially observed settings (POMDPs).

ST-DIM \citep{anand2019unsupervised} introduced various temporal, contrastive losses, including ones that operate on ``local'' features from an intermediate layer within the encoder, without data augmentation.  CURL \citep{laskin2020curl} introduced an augmented, contrastive auxiliary task similar to ours, including a momentum encoder but without temporal contrast. \cite{mazoure2020deep} provided extensive analysis pertaining to InfoNCE losses on functions of successive time steps in MDPs, including local features in their auxiliary loss (DRIML) similar to ST-DIM, and finally conducted experiments using global temporal contrast of augmented observations in the Procgen \citep{cobbe2019procgen} environment. Most recently, MPR \citep{schwarzer2020} combined data augmentation with multi-step, convolutional forward modeling and a similarity loss to improve DQN agents in the Atari 100k benchmark.  
\cite{hafner2018learning,hafner2019dream,lee2019stochastic} proposed to leverage world-modeling in a latent-space for continuous control.
A small number of model-free methods have attempted to decouple encoder training from the RL loss as ablations, but have met reduced performance relative to end-to-end RL \citep{laskin2020curl,lee2020predictive}.
None have previously been shown effective in as diverse a collection of RL environments as ours \citep{bellemare2013arcade,tassa2018deepmind,dmlab}.

\cite{finn_spationalautoenc,ha2018world} are example works which pretrained encoder features in advance using image reconstruction losses such as the VAE \citep{kingma2013auto}.  \cite{devin2018deep,kipf2019contrastive} pretrained object-centric representations, the latter  learning a forward model by way of contrastive losses; \cite{yan2020learning} introduced a similar technique to learn encoders supporting manipulation of deformable objects by traditional control methods.  MERLIN \citep{wayne2018unsupervised} trained a convolutional encoder and sophisticated memory module online, detached from the RL agent, which learned read-only accesses to memory.  It used reconstruction and one-step latent-prediction losses and achieved high performance in DMLab-like environments with extreme partial observability.  Our loss function may benefit those settings, as it outperforms similar reconstruction losses in our experiments.  Decoupling unsupervised pretraining from downstream tasks is common in computer vision \citep{henaff2019data,kaiming2019moco,chen2020simclr} and has favorable properties of providing task agnostic features which can be used for training smaller task-specific networks, yielding significant gains in computational efficiency over end-to-end methods.  

\begin{figure}[h]
    \centering
    \includegraphics[width=0.39\textwidth]{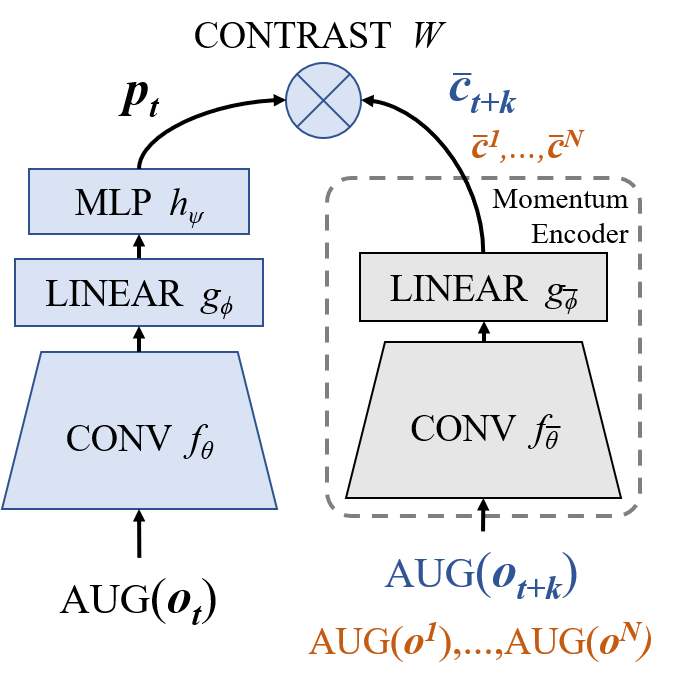}
  \caption{\small Augmented Temporal Contrast---augmented observations are processed through a learned encoder $f_\theta$, compressor, $g_\phi$ and residual predictor $h_\psi$, and are associated through a contrastive loss with a positive example from $k$ time steps later, processed through a momentum encoder.}
  \label{fig:atc}
\end{figure}

\section{Augmented Temporal Contrast}  
Our unsupervised learning task, Augmented Temporal Contrast (ATC), requires a model to associate an observation, $o_t$, with one from a specified, near-future time step, $o_{t+k}$.  Within each training batch, we apply stochastic data augmentation to the observations \citep{laskin2020curl}, namely random shift \citep{kostrikov2020image}, which is simple to implement and provides highly effective regularization in most cases.  The augmented observations are encoded into a small latent space where a contrastive loss is applied.  This task encourages the learned encoder to extract meaningful elements of the structure of the MDP from observations.


Our architecture for ATC consists of four learned components - (i) a convolutional \emph{encoder}, $f_\theta$, which processes the anchor observation, $o_t$, into the latent image $z_t=f_\theta(\textsc{aug}(o_t))$, (ii) a linear \emph{global compressor}, $g_\phi$ to produce a small latent code vector $c_t=g_\phi(z_t)$, (iii) a residual \emph{predictor} MLP, $h_\psi$, which acts as an implicit forward model to advance the code $p_t=h_\psi(c_t)+c_t$, and (iv) a \emph{contrastive transformation} matrix, $W$.  To process the positive observation, $o_{t+k}$ into the target code $\bar{c}_{t+k}=g_{\bar{\phi}} (f_{\bar{\theta}} (\textsc{aug}(o_{t+k}))$, we use a momentum encoder \citep{kaiming2019moco} parameterized as a slowly moving average of the weights from the learned encoder and compressor layer:
\begin{equation}
    \bar{\theta}\gets (1-\tau)\bar{\theta} + \tau\theta\ \ ;\qquad \bar{\phi}\gets (1-\tau)\bar{\phi} + \tau\phi \ .
\end{equation}
The complete architecture is shown in Figure \ref{fig:atc}.  The convolutional encoder, $f_\theta$, alone is shared with the RL agent.  

We employ the InfoNCE loss \citep{infonce,oord2018representation} using logits computed bilinearly, as $l=p_tW\bar{c}_{t+k}$.  
In our implementation, every anchor in the training batch utilizes the positives corresponding to all other anchors as its negative examples. Denoting an observation indexed from dataset $\mathcal{O}$ as $o_i$, and its positive as $o_{i+}$, the logits can be written as $l_{i,j+}=p_i W\bar{c}_{j+}$; our loss function in practice is:
\begin{equation}
    \mathcal{L}^{ATC}=-\mathbb{E}_\mathcal{O}\left[\log \frac{\exp{l_{i,i+}}}{\sum_{o_j\in \mathcal{O}}\exp{ l_{i,j+}}}    \right].
\end{equation}
 
\section{Experiments}

\subsection{Evaluation Environments and Algorithms}
We evaluate ATC on three standard, visually diverse RL benchmarks - the DeepMind control suite (DMControl; \citealt{tassa2018deepmind}), Atari games in the Arcade Learning Environment \citep{bellemare2013arcade}, and DeepMind Lab (DMLab; \citealt{dmlab}). Atari requires discrete control in arcade-style games.  DMControl is comprised of continuous control robotic locomotion and manipulation tasks.  In contrast, DMLab requries the RL agent to reason in more visually complex 3D maze environments with partial observability.

We use ATC to enhance both on-policy and off-policy RL algorithms. For DMControl, we use RAD-SAC \citep{laskin_lee2020rad,haarnoja2018soft} with the augmentation of \cite{kostrikov2020image}, which randomly shifts the image in each coordinate (by up to 4 pixels), replicating edge pixel values as necessary to restore the original image size.  A difference from prior work is that we use more downsampling in our convolutional network, by using strides $(2, 2, 2, 1)$ instead of $(2, 1, 1, 1)$ to reduce the convolution output image by 25x.\footnote{For our input image size $84\times84$, the convolution output image is $7\times7$ rather than $35\times35$.  Performance remains largely unchanged, except for a small decrease in the \textsc{half-cheetah} environment, but the experiments run significantly faster and use less GPU memory.}  For both Atari and DMLab, we use PPO \citep{schulman2017proximal}.  In Atari, we use feed-forward agents, sticky actions, and no end-of-life boundaries for RL episodes.  In DMLab we used recurrent, LSTM agents receiving only a single time-step image input, the four-layer convolution encoder from \cite{jaderberg2019human}, and we tuned the entropy bonus for each level.  In the online setting, the ATC loss is trained using small replay buffer of recent experiences.  

We include all our own baselines for fair comparison and provide  complete settings in an appendix.  In each plot, the bold lines show the average return across seeds, and the shaded area around each curve represents one standard deviation in measured return.  

\subsection{Online RL with ATC}

\begin{figure*}[h]
    \centering
    \includegraphics[width=\textwidth]{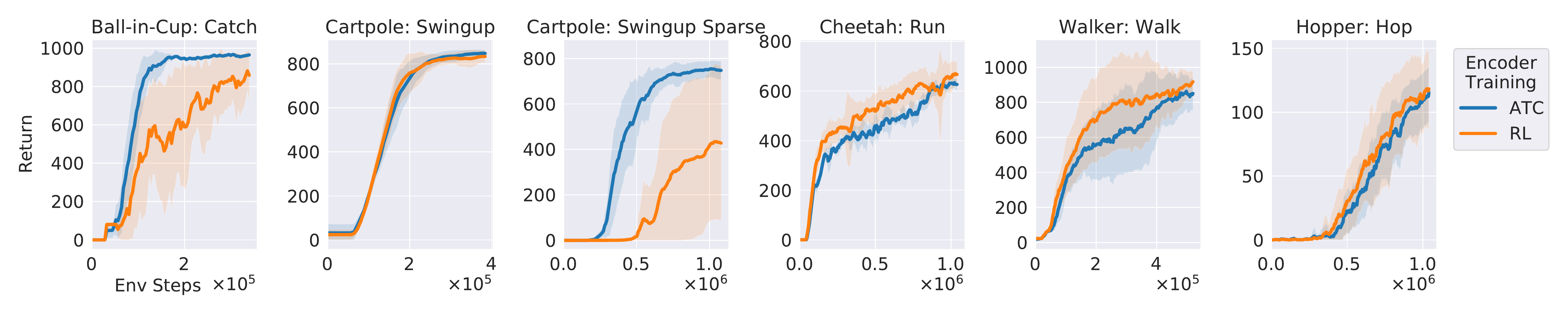}
    \vskip -0.1in
    \caption{\small Online encoder training by ATC, fully detached from RL training, performs as well as end-to-end RL in DMControl, and better in sparse-reward environments (environment steps shown, see appendix for action repeats). Each curve is 10 random seeds.}
    \label{fig:dmc_sac_online}
\end{figure*}

\paragraph{DMControl}  
In the online setting, we found ATC to be capable of training the encoder by itself (\textit{i.e.}, with encoder fully detached from any RL gradient update), achieving essentially equal or better scores versus end-to-end RL in all six environments we tested, Figure~\ref{fig:dmc_sac_online}.  In \textsc{Cartpole-Swingup-Sparse}, where rewards are only received once the pole reaches vertical, ATC training enabled the agent to master the task significantly faster.  The encoder is trained with one update for every RL update to the policy, using the same batch size, except in \textsc{Cheetah-Run}, which required twice the rate of ATC
updates.

\begin{figure*}[h]
    \centering
    \includegraphics[width=0.95\textwidth]{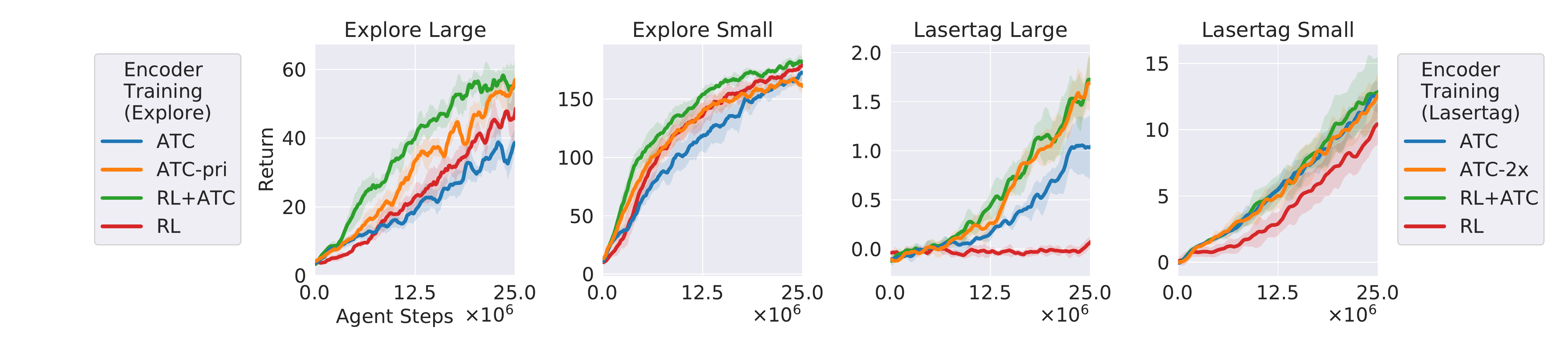}
    \vskip -0.1in
    \caption{\small Online encoder training by ATC, fully detached from the RL agent, performs as well or better than end-to-end RL in DMLab (1 agent step = 4 environment steps, the standard action repeat).
    Prioritized ATC replay (\textsc{Explore}) or increased ATC training (\textsc{Lasertag}) addresses sparsities to nearly match performance of RL with ATC as an auxiliary loss (RL+ATC). Each curve is 5 random seeds.}
    \label{fig:dmlab_ppo_online}
\end{figure*}

\paragraph{DMLab}  We experimented with two kinds of levels in DMLab: \textsc{Explore\_Goal\_Locations}, which requires repeatedly navigating a maze whose layout is randomized every episode, and \textsc{Lasertag\_Three\_Opponents}, which requires fast reflexes to pursue and tag enemies at a distance.   We found ATC capable of training fully detached encoders while achieving equal or better performance than end-to-end RL.  Results are shown in Figure~\ref{fig:dmlab_ppo_online}.  Both environments exhibit sparsity which is greater in the ``large'' version than the ``small'' version, which our algorithm addresses, discussed next.

In \textsc{Explore}, the goal object is rarely seen, especially early on, making its appearance difficult to learn.  We therefore introduced prioritized sampling for ATC, with priorities corresponding to empirical absolute returns: $p \propto 1 + R_{abs}$, where $R_{abs}=\sum_{t=0}^n \gamma^t |r_t|$, to train more frequently on more informative scenes.\footnote{In \textsc{Explore\_Goal\_Locations}, the only reward is +10, earned when reaching the goal object.}  Whereas uniform-ATC performs slightly below RL, prioritized-ATC outperforms RL and nearly matches using ATC (uniform) as an auxiliary task.  By considering the encoder as a stand-alone feature extractor separate from the policy, no importance sampling correction is required.  

In \textsc{Lasertag}, enemies are often seen, but the reward of tagging one is rarely achieved by the random agent.  ATC learns the relevant features anyway, boosting performance while the RL-only agent remains at zero average score.  We found that increasing the rate of UL training to do twice as many updates\footnote{Since the ATC batch size was 512 but the RL batch size was 1024, performing twice as many UL updates still only consumed the same amount of encoder training data as RL.  We did not fine-tune for batch size.} further improved the score to match the ATC-auxiliary agent, showing flexibility to address the representation-learning bottleneck when opponents are dispersed.

\begin{figure*}[h]
    \centering
    \includegraphics[width=0.95\textwidth]{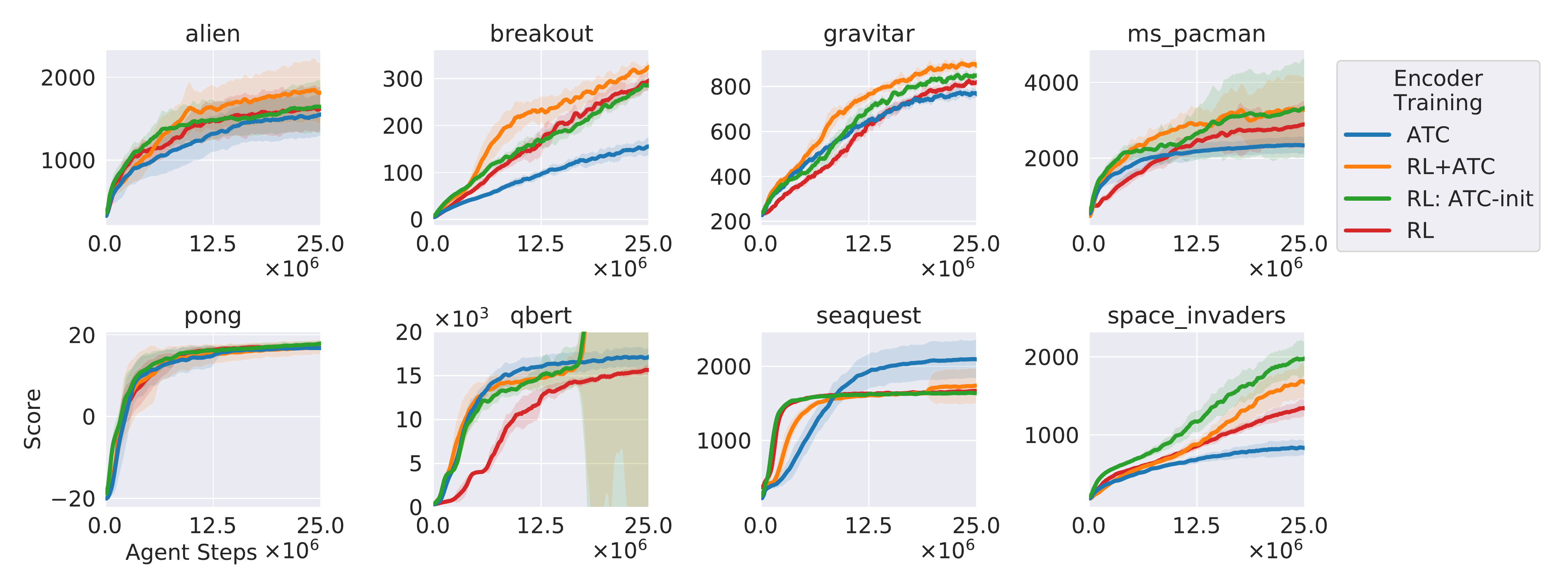}
    \vskip -0.1in
    \caption{\small Online encoder training using ATC, fully detached from the RL agent, works well in 5 of 8 games tested (1 agent step = 4 environment steps, the standard action repeat).  6 of 8 games benefit significantly from using ATC as an auxiliary loss or for weight initialization. Each curve is 8 random seeds.}
    \label{fig:atari_ppo_online}
\end{figure*}

\paragraph{Atari}  We tested a diverse subset of eight Atari games, shown in Figure~\ref{fig:atari_ppo_online}.  We found detached-encoder training to work as well as end-to-end RL in five games, but performance suffered in \textsc{Breakout} and \textsc{Space Invaders} in particular.  Using ATC as an auxiliary task, however, improves performance in these games and others.  We found it helpful to anneal the amount of UL training over the course of RL in Atari (details in an appendix).  Notably, we found several games, including \textsc{Space Invaders}, to benefit from using ATC only to initialize encoder weights, done using an initial 100k transitions gathered with a uniform random policy.  Some of our remaining experiments provide more insights into the challenges of this domain.

\subsection{Encoder Pre-Training Benchmarks}
To benchmark the effectiveness of different UL algorithms for RL, we propose a new evaluation methodology that is similar to how UL pre-training techniques are measured in computer vision (see \textit{e.g.} \cite{chen2020simclr,grill2020bootstrap}): 
(i) collect a data set composed of expert demonstrations from each environment; (ii) pre-train the CNN encoder with that data offline using UL; (iii) evaluate by using RL to learn a control policy while keeping the encoder weights frozen.  This procedure isolates the asymptotic performance of each UL algorithm for RL.  For convenience, we drew expert demonstrations from partially-trained RL agents, and every UL algorithm trained on the same data set for each environment.  Our RL agents used the same post-encoder architectures as in the online experiments.  Further details about pre-training by each algorithm are provided in an appendix.

\begin{figure*}[h]
    \centering
    \includegraphics[width=0.95\textwidth]{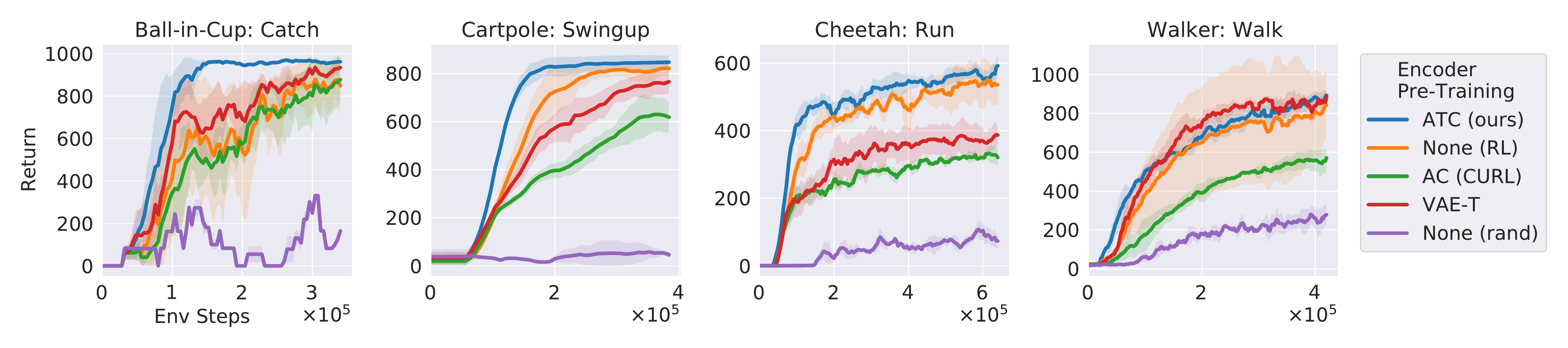}
    \vskip -0.1in
    \caption{\small RL in DMControl, using encoders pre-trained on expert demonstrations using UL, with weights frozen---across all domains, ATC outperforms prior methods and the end-to-end RL reference. Each curve is a mininum of 4 random seeds.}
    \label{fig:dmc_sac_offline}
    \vspace{-3mm}
\end{figure*}

\paragraph{DMControl}  We compare ATC against two competing algorithms: Augmented Contrast (AC), from CURL \citep{laskin2020curl}, which uses the same observation for the anchor and the positive, and a VAE \citep{kingma2013auto}, for which we found better performance by introducing a time delay to the target observation (VAE-T).  We found ATC to match or outperform the other algorithms, in all four test environments, as shown in Figure~\ref{fig:dmc_sac_offline}.  Further, ATC is the only one to match or outperform the reference end-to-end RL across all cases.

\begin{figure}[h]
    \centering
    \includegraphics[width=0.45\textwidth]{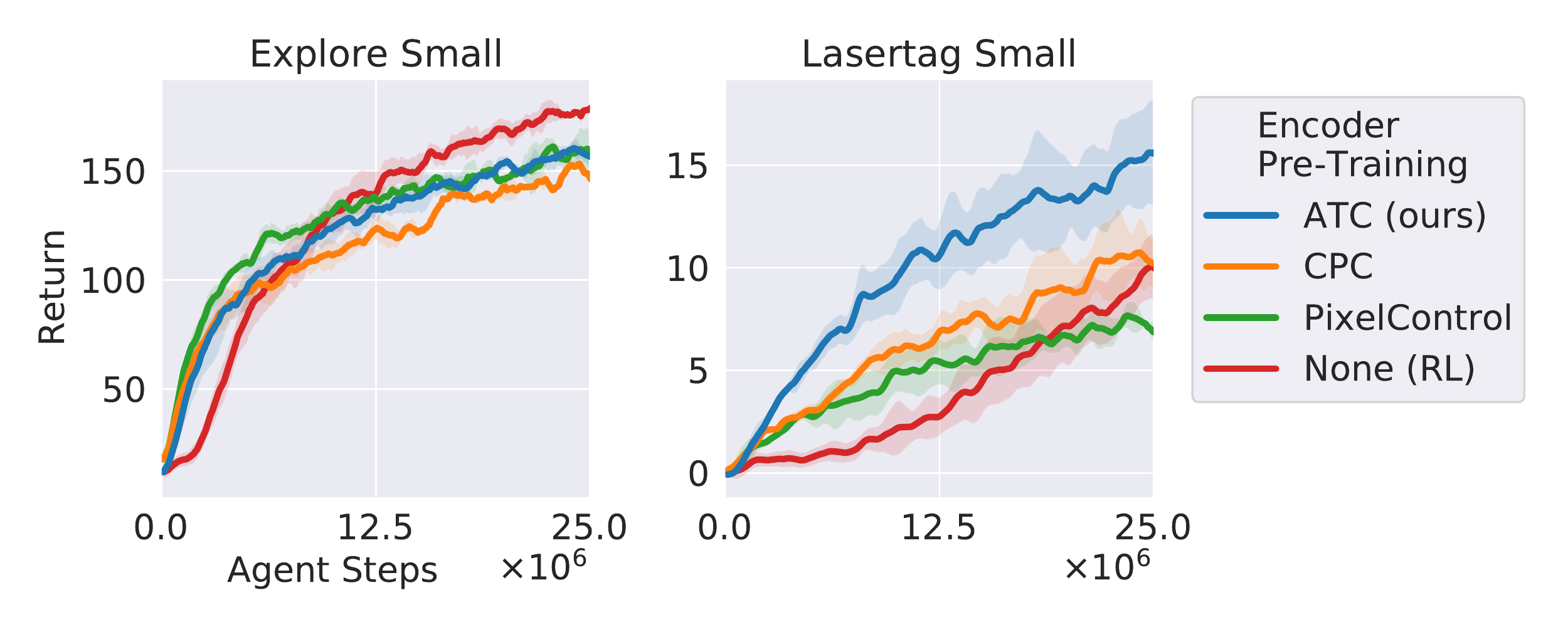}
    \vskip -0.1in
  \caption{\small RL in DMLab, using pre-trained encoders with weights frozen--in \textsc{Lasertag} especially, ATC outperforms leading prior UL algorithms. Each curve is 4 random seeds.}
  \label{fig:dmlab_ppo_offline}
  \vspace{-4mm}
\end{figure}

\paragraph{DMLab}  We compare against both Pixel Control \citep{jaderberg2016reinforcement,hessel2019multi} and CPC \citep{oord2018representation}, which have been shown to bring strong benefits in DMLab.  While all algorithms perform similarly well in \textsc{Explore}, ATC performs significantly better in \textsc{Lasertag}, Figure~\ref{fig:dmlab_ppo_offline}.  Our algorithm is simpler than Pixel Control and CPC in the sense that it uses neither actions, deconvolution, nor recurrence.  

\begin{figure*}[h]
    \centering
    \includegraphics[width=0.9\textwidth]{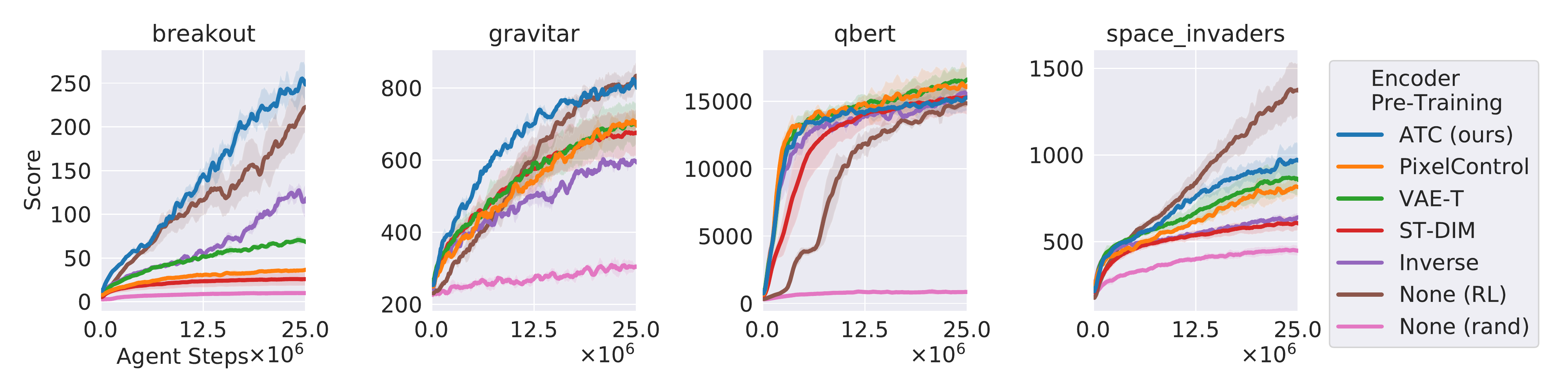}
    \vskip -0.1in
    \caption{\small RL in Atari, using pre-trained encoders with weights frozen---ATC outperforms several leading, prior UL algorithms and exceeds the end-to-end RL reference in 3 of the 4 games tested.  Each curve is minimum 3 random seeds.}
    \label{fig:atari_ppo_offline}
\end{figure*}

\paragraph{Atari}  We compare against Pixel Control, VAE-T, and a basic inverse model which predicts actions between pairs of observations.  We also compare against Spatio-Temporal Deep InfoMax (ST-DIM), which uses temporal contrastive losses with ``local'' features from an intermediate convolution layer to ensure attention to the whole screen; it was shown to produce detailed game-state knowledge when applied to individual frames \citep{anand2019unsupervised}.  Of the four games shown in Figure~\ref{fig:atari_ppo_offline}, ATC is the only UL algorithm to match the end-to-end RL reference in \textsc{Gravitar} and \textsc{Breakout}, and it performs best in \textsc{Space Invaders}.

\subsection{Multi-Task Encoders}

\begin{figure*}[h]
    \centering
    \includegraphics[width=0.9\textwidth]{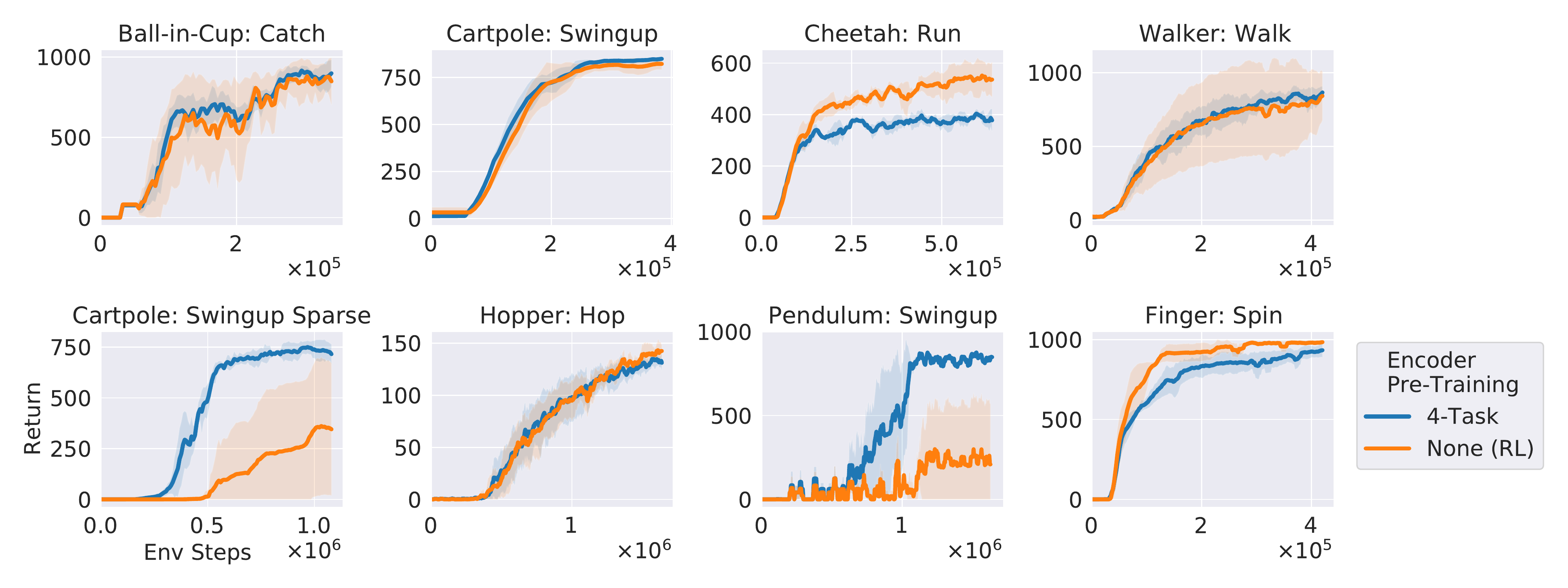}
    \vskip -0.1in
    \caption{\small Separate RL agents using a single encoder with weights frozen after pre-training on expert demonstrations from the four top environments.   The encoder generalizes to four new environments, bottom row, where sparse reward tasks especially benefit from the transfer. Each curve is minimum 4 random seeds.}
    \label{fig:dmc_sac_multi}
\end{figure*}

In the offline setting, we conducted initial explorations into the capability of ATC to learn multi-task encoders, simply by pre-training on demonstrations from multiple environments.  We evaluate the encoder by using it, with frozen weights, in separate RL agents learning each downstream task.  
\paragraph{DMControl} Figure ~\ref{fig:dmc_sac_multi} shows our results in DMControl, where we pretrained using only the four environments in the top row.  Although the encoder was never trained on the \textsc{Hopper}, \textsc{Pendulum}, nor \textsc{Finger} domains, the multi-task encoder supports efficient RL in them.  \textsc{Pendulum-Swingup} and \textsc{Cartpole-Swingup-Sparse} stand out as challenging environments which benefited from cross-domain and cross-task pre-training, respectively.  The pretraining was remarkably efficient, requiring only 20,000 updates to the encoder.

\paragraph{Atari}  Atari proved a more challenging domain for learning multi-task encoders.  Learning  all eight games together in Figure~\ref{fig:atari_ppo_multi}, in the appendix, resulted in diminished performance relative to single-game pretraining in three of the eight.  The decrease was partially alleviated by widening the encoder with twice as many filters per layer, indicating that representation capacity is a limiting factor.  To test generalization, we conducted a seven-game pre-training experiment where we test the encoder on the held-out game.  Most games suffered diminished performance (although still perform significantly higher than a frozen random encoder), confirming the limited extent to which visual features transfer across these games.

\subsection{Ablations and Encoder Analysis}

\begin{figure*}[h]
    \centering
    \includegraphics[width=0.75\textwidth]{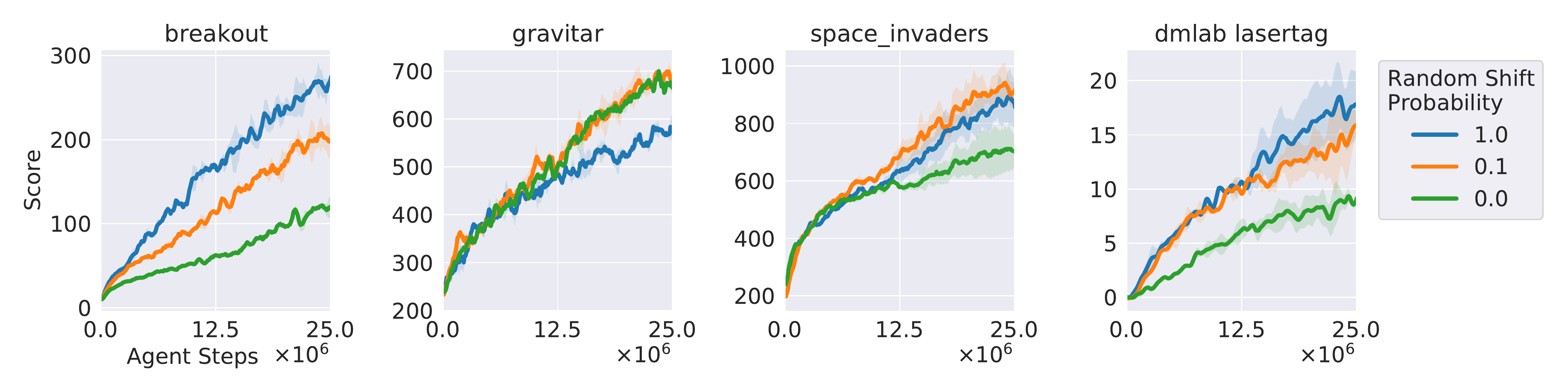}
    \caption{\small Random shift augmentation helps in some Atari games and hurts in others, but applying to each image with probability 0.1 is a performant middle ground.  DMLab, a 3-D visual environment, benefits from random shift. (Offline pre-training.)}
    \label{fig:atari_ppo_RS_ablation}
\end{figure*}

\paragraph{Random Shift in ATC}  In offline experiments, we discovered random shift augmentations to be helpful in all domains.  To our knowledge, this is the first application of random shift to 3D visual environments as in DMLab.  In Atari, we found performance in \textsc{Gravitar} to suffer from random shift, but reducing the probability of applying random shift to each observation from 1.0 to 0.1 alleviated the effect while still bringing benefits in other games, so we used this setting in our main experiments.  Results are shown in Figure~\ref{fig:atari_ppo_RS_ablation}.
\vspace{-4mm}

\paragraph{Random Shift in RL}  In DMControl, we found the best results when using random shift during RL, even when training with a frozen encoder.  This is evidence that the augmentation regularizes not only the representation but also the policy, which first processes the latent image into a 50-dimensional vector.  To unlock computation and memory benefits of replaying only the latent images for the RL agent, we attempted to apply data augmentation to the latent image.  But we found the smallest possible random shifts to be too extreme.  Instead, we introduce a new augmentation, \emph{subpixel random shift}, which linearly interpolates among neighboring pixels.  

\begin{figure*}[h]
    \centering
    \includegraphics[width=0.75\textwidth]{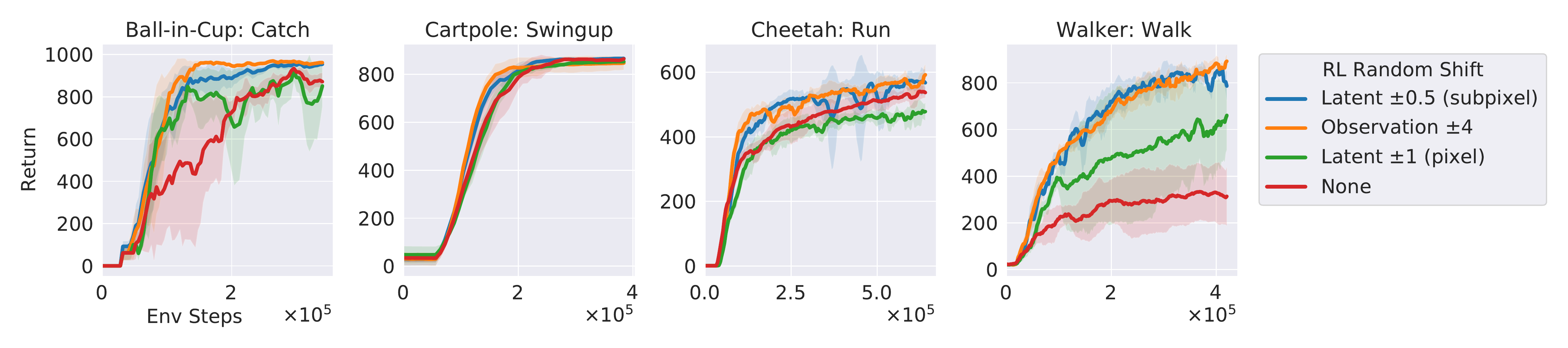}
    \caption{\small Even after pre-training encoders for DMControl using random shift, RL requires augmentation---our subpixel augmentation acts on the (compressed) latent image, permitting its use in the replay buffer.}
    \label{fig:dmc_sac_subpixel}
    
\end{figure*}

\begin{figure}
    \centering
    \includegraphics[width=0.3\textwidth]{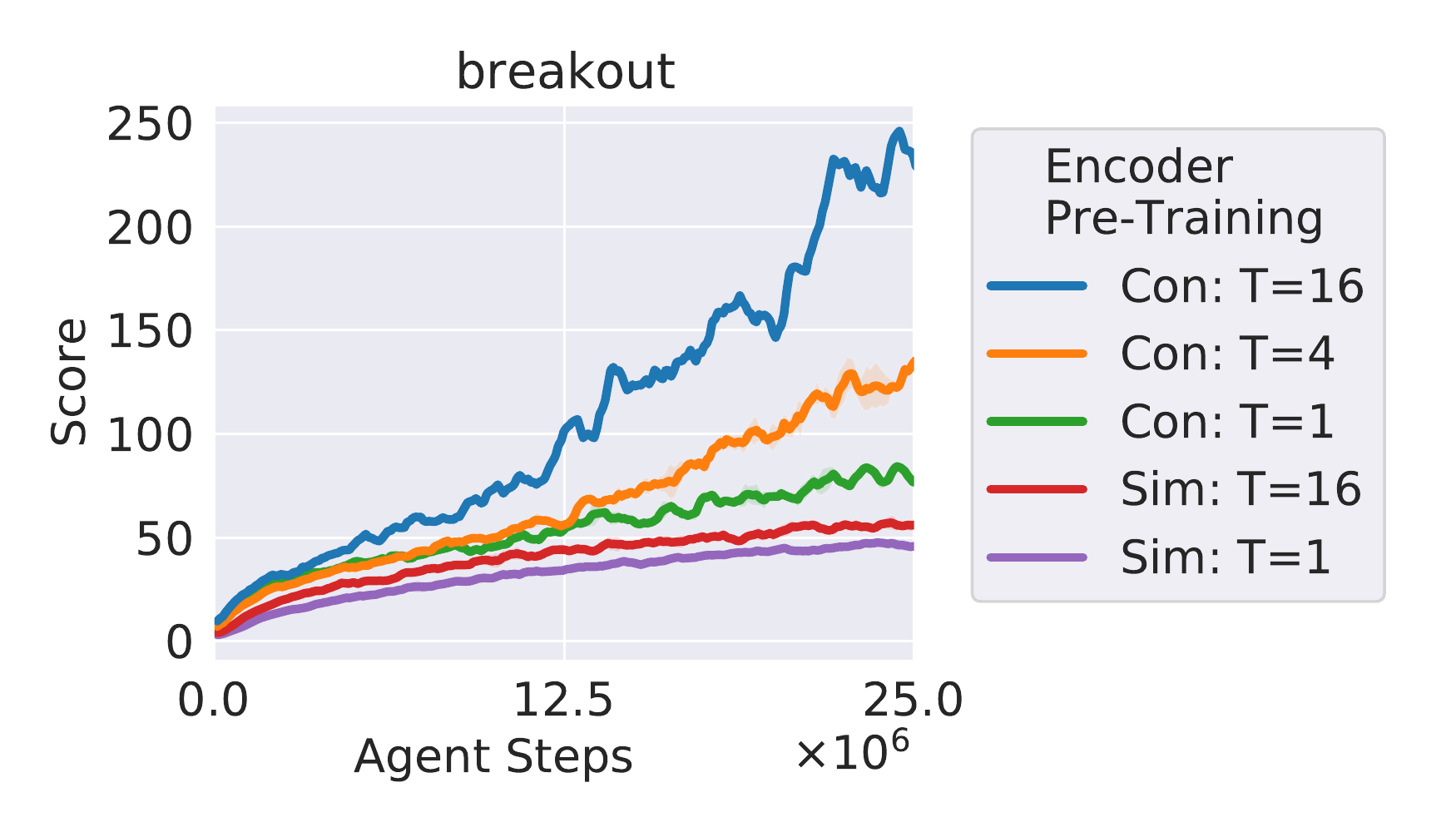} 
    \vspace{-0.2in}
    \caption{\small \textsc{Breakout} benefits from contrasting against negatives from several neighboring time steps.}
    \label{fig:atari_ppo_TB_ablation}
\end{figure}

In subpixel random shift, new pixels are a linearly weighted average of the four nearest pixels to a randomly chosen coordinate location.  We used uniformly random horizontal and vertical shifts, and tested maximum displacements in $(\pm)\ \{0.1, 0.25, 0.5, 0.75, 1.0\}$ pixels (with ``edge'' mode padding $\pm1$).  We found $0.5$ to work well in all tested domains, restoring the performance of raw image augmentation but eliminating convolutions entirely from the RL training updates. 
As shown in Figure~\ref{fig:dmc_sac_subpixel}, this augmentation restores performance when applied to the latent images, allowing a pre-trained encoder to be entirely bypassed during policy training updates.
\vspace{-2mm}
\paragraph{Temporal Contrast on Sequences}  In \textsc{Breakout} alone, we discovered that composing the UL training batch of trajectory segments, rather than individual transitions, gave a significant benefit.  Treating all elements of the training batch independently provides ``hard'' negatives, since the encoder must distinguish between neighboring time steps.  This setting had no effect in the other Atari games tested, and we found equal or better performance using individual transitions in DMControl and DMLab.  Figure~\ref{fig:atari_ppo_TB_ablation} further shows that using a similarity loss \citep{grill2020bootstrap} does not capture the benefit.
\vspace{-3mm}

\begin{figure}
    \centering
    \includegraphics[width=0.48\textwidth]{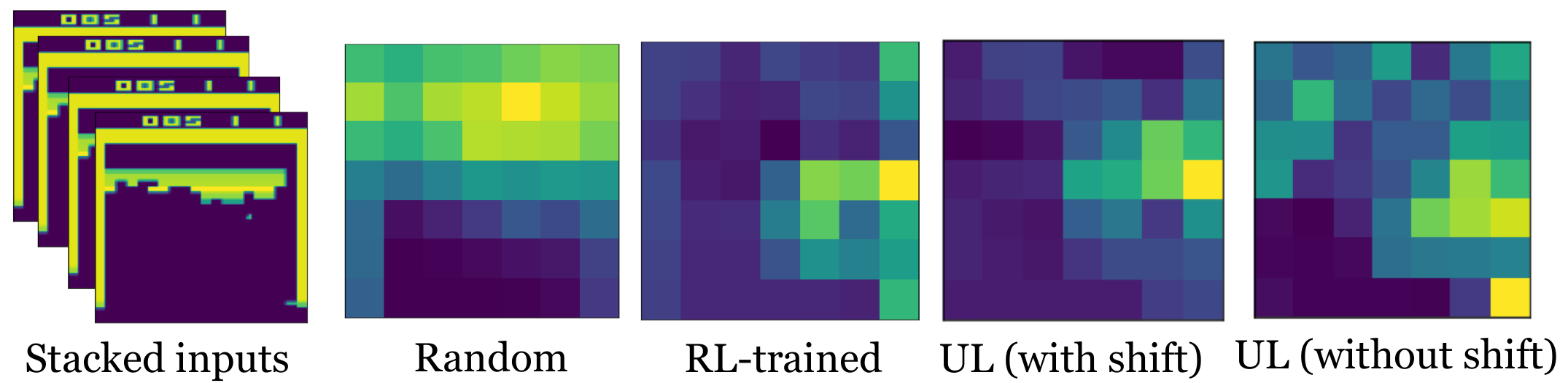} 
    \vspace{-4mm}
    \caption{\small An example scene from \textsc{Breakout}, where a low-performance UL encoder (without shift) focuses on the paddle.  Introducing random shift and sequence data makes the high-performance UL encoder (full ATC) focus near the ball, as does the encoder from a fully-trained, end-to-end RL agent.}
    \label{fig:breakout_attention_2}
    \vspace{-6mm}
\end{figure}

\paragraph{Encoder Analysis}  We analyzed the learned encoders in \textsc{Breakout} to further study this ablation effect. Similar to \cite{zagoruyko2016paying}, we compute spatial attention maps by mean-pooling the absolute values of the activations along the channel dimension and follow with a 2-dimensional spatial softmax. Figure~\ref{fig:breakout_attention_2} shows the attention of four different encoders on the displayed scene.  The poorly performing UL encoder heavily utilizes the paddle to distinguish the observation.  The UL encoder trained with random shift and sequence data, however, focuses near the ball, as does the fully-trained RL encoder.  (The random encoder mostly highlights the bricks, which are less relevant for control.)  In an appendix, we include other example encoder analyses from Atari and DMLab which show ATC-trained encoders attending only to key objects on the game screen, while RL-trained encoders additionally attend to potentially distracting features such as game score.

We note anecdotally that most of our ATC encoders trained offline reached an accuracy well in excess of 90\% at identifying the correct positive example among the hundreds of negatives present in the training batch.  Applying random shift augmentation often reduced the accuracy, even while improving RL performance.  Among the environments we tested, we did not observe a strong correlation between ATC accuracy at convergence and downstream RL performance relative to end-to-end training, suggesting more research could improve understanding of feature-relevance for RL.

\section{Conclusion}
Reward-free representation learning from images provides flexibility and insights for improving deep RL agents.  We have shown a broad range of cases where our new unsupervised learning algorithm can fully replace RL for training convolutional encoders while maintaining or improving online performance.  In a small number of environments--a few of the Atari games--including the RL loss for encoder training still surpasses our UL-only method, leaving opportunities for further improvements in UL for RL.  

Our preliminary efforts to use actions as inputs (into the \emph{predictor} MLP) or as prediction outputs (inverse loss) with ATC did not immediately yield strong improvements.  We experimented only with random shift, but other augmentations may be useful, as well.  In multi-task encoder training, our technique avoids any need for sophisticated reward-balancing \citep{hessel2019multi}, but more advanced training methods may still help when the required features are in conflict, as in Atari, or if they otherwise impact our loss function unequally.  On the theoretical side, it may be helpful to analyze the effects of domain shift on the policy when a detached representation is learned online.

One obvious application of our offline methodology would be in the batch RL setting, where the agent learns from a fixed data set.  Our offline experiments showed that a relatively small number of transitions are sufficient to learn rich representations by UL, and the lower limit could be further explored.  Overall, we hope that our algorithm and experiments spur further developments leveraging unsupervised learning for reinforcement learning.
\vspace{-4mm}


\section*{Acknowledgments} This work was supported by ONR PECASE N00014161723 and Open Philanthropy. We thank Ankesh Anand and Evan Racah for helpful discussions.

\bibliography{main}

\begin{thebibliography}{8}
\providecommand{\natexlab}[1]{#1}
\providecommand{\url}[1]{\texttt{#1}}
\expandafter\ifx\csname urlstyle\endcsname\relax
  \providecommand{\doi}[1]{doi: #1}\else
  \providecommand{\doi}{doi: \begingroup \urlstyle{rm}\Url}\fi

\bibitem[Author(2021)]{anonymous}
Author, N.~N.
\newblock Suppressed for anonymity, 2021.

\bibitem[Duda et~al.(2000)Duda, Hart, and Stork]{DudaHart2nd}
Duda, R.~O., Hart, P.~E., and Stork, D.~G.
\newblock \emph{Pattern Classification}.
\newblock John Wiley and Sons, 2nd edition, 2000.

\bibitem[Kearns(1989)]{kearns89}
Kearns, M.~J.
\newblock \emph{Computational Complexity of Machine Learning}.
\newblock PhD thesis, Department of Computer Science, Harvard University, 1989.

\bibitem[Langley(2000)]{langley00}
Langley, P.
\newblock Crafting papers on machine learning.
\newblock In Langley, P. (ed.), \emph{Proceedings of the 17th International
  Conference on Machine Learning (ICML 2000)}, pp.\  1207--1216, Stanford, CA,
  2000. Morgan Kaufmann.

\bibitem[Michalski et~al.(1983)Michalski, Carbonell, and
  Mitchell]{MachineLearningI}
Michalski, R.~S., Carbonell, J.~G., and Mitchell, T.~M. (eds.).
\newblock \emph{Machine Learning: An Artificial Intelligence Approach, Vol. I}.
\newblock Tioga, Palo Alto, CA, 1983.

\bibitem[Mitchell(1980)]{mitchell80}
Mitchell, T.~M.
\newblock The need for biases in learning generalizations.
\newblock Technical report, Computer Science Department, Rutgers University,
  New Brunswick, MA, 1980.

\bibitem[Newell \& Rosenbloom(1981)Newell and Rosenbloom]{Newell81}
Newell, A. and Rosenbloom, P.~S.
\newblock Mechanisms of skill acquisition and the law of practice.
\newblock In Anderson, J.~R. (ed.), \emph{Cognitive Skills and Their
  Acquisition}, chapter~1, pp.\  1--51. Lawrence Erlbaum Associates, Inc.,
  Hillsdale, NJ, 1981.

\bibitem[Samuel(1959)]{Samuel59}
Samuel, A.~L.
\newblock Some studies in machine learning using the game of checkers.
\newblock \emph{IBM Journal of Research and Development}, 3\penalty0
  (3):\penalty0 211--229, 1959.

\end{thebibliography}


\begin{thebibliography}{43}
\providecommand{\natexlab}[1]{#1}
\providecommand{\url}[1]{\texttt{#1}}
\expandafter\ifx\csname urlstyle\endcsname\relax
  \providecommand{\doi}[1]{doi: #1}\else
  \providecommand{\doi}{doi: \begingroup \urlstyle{rm}\Url}\fi

\bibitem[Anand et~al.(2019)Anand, Racah, Ozair, Bengio, C{\^o}t{\'e}, and
  Hjelm]{anand2019unsupervised}
Anand, A., Racah, E., Ozair, S., Bengio, Y., C{\^o}t{\'e}, M.-A., and Hjelm,
  R.~D.
\newblock Unsupervised state representation learning in atari.
\newblock In \emph{Advances in Neural Information Processing Systems}, 2019.

\bibitem[Beattie et~al.(2016)Beattie, Leibo, Teplyashin, Ward, Wainwright,
  Küttler, Lefrancq, Green, Valdés, Sadik, Schrittwieser, Anderson, York,
  Cant, Cain, Bolton, Gaffney, King, Hassabis, Legg, and Petersen]{dmlab}
Beattie, C., Leibo, J.~Z., Teplyashin, D., Ward, T., Wainwright, M., Küttler,
  H., Lefrancq, A., Green, S., Valdés, V., Sadik, A., Schrittwieser, J.,
  Anderson, K., York, S., Cant, M., Cain, A., Bolton, A., Gaffney, S., King,
  H., Hassabis, D., Legg, S., and Petersen, S.
\newblock Deepmind lab.
\newblock \emph{arXiv preprint arXiv:1612.03801}, 2016.

\bibitem[Bellemare et~al.(2013)Bellemare, Naddaf, Veness, and
  Bowling]{bellemare2013arcade}
Bellemare, M.~G., Naddaf, Y., Veness, J., and Bowling, M.
\newblock The arcade learning environment: An evaluation platform for general
  agents.
\newblock \emph{Journal of Artificial Intelligence Research}, 47:\penalty0
  253--279, 2013.

\bibitem[Berner et~al.(2019)Berner, Brockman, Chan, Cheung, Debiak, Dennison,
  Farhi, Fischer, Hashme, Hesse, et~al.]{openai2019dota}
Berner, C., Brockman, G., Chan, B., Cheung, V., Debiak, P., Dennison, C.,
  Farhi, D., Fischer, Q., Hashme, S., Hesse, C., et~al.
\newblock Dota 2 with large scale deep reinforcement learning.
\newblock \emph{arXiv preprint arXiv:1912.06680}, 2019.

\bibitem[Chen et~al.(2020)Chen, Kornblith, Norouzi, and Hinton]{chen2020simclr}
Chen, T., Kornblith, S., Norouzi, M., and Hinton, G.
\newblock A simple framework for contrastive learning of visual
  representations.
\newblock \emph{arXiv:2002.05709}, 2020.

\bibitem[Cobbe et~al.(2019)Cobbe, Hesse, Hilton, and
  Schulman]{cobbe2019procgen}
Cobbe, K., Hesse, C., Hilton, J., and Schulman, J.
\newblock Leveraging procedural generation to benchmark reinforcement learning.
\newblock \emph{arXiv preprint arXiv:1912.01588}, 2019.

\bibitem[Devin et~al.(2018)Devin, Abbeel, Darrell, and Levine]{devin2018deep}
Devin, C., Abbeel, P., Darrell, T., and Levine, S.
\newblock Deep object-centric representations for generalizable robot learning.
\newblock In \emph{2018 IEEE International Conference on Robotics and
  Automation (ICRA)}, pp.\  7111--7118. IEEE, 2018.

\bibitem[Dosovitskiy et~al.(2017)Dosovitskiy, Ros, Codevilla, Lopez, and
  Koltun]{dosovitskiy2017}
Dosovitskiy, A., Ros, G., Codevilla, F., Lopez, A., and Koltun, V.
\newblock Carla: An open urban driving simulator.
\newblock \emph{arXiv preprint arXiv:1711.03938}, 2017.

\bibitem[{Finn} et~al.(2016){Finn}, {Xin Yu Tan}, {Yan Duan}, {Darrell},
  {Levine}, and {Abbeel}]{finn_spationalautoenc}
{Finn}, C., {Xin Yu Tan}, {Yan Duan}, {Darrell}, T., {Levine}, S., and
  {Abbeel}, P.
\newblock Deep spatial autoencoders for visuomotor learning.
\newblock In \emph{2016 IEEE International Conference on Robotics and
  Automation (ICRA)}, pp.\  512--519, 2016.

\bibitem[Grill et~al.(2020)Grill, Strub, Altch{\'e}, Tallec, Richemond,
  Buchatskaya, Doersch, Pires, Guo, Azar, et~al.]{grill2020bootstrap}
Grill, J.-B., Strub, F., Altch{\'e}, F., Tallec, C., Richemond, P.~H.,
  Buchatskaya, E., Doersch, C., Pires, B.~A., Guo, Z.~D., Azar, M.~G., et~al.
\newblock Bootstrap your own latent: A new approach to self-supervised
  learning.
\newblock \emph{arXiv preprint arXiv:2006.07733}, 2020.

\bibitem[Guo et~al.(2020)Guo, Pires, Piot, Grill, Altch{\'e}, Munos, and
  Azar]{guo2020bootstrap}
Guo, D., Pires, B.~A., Piot, B., Grill, J.-b., Altch{\'e}, F., Munos, R., and
  Azar, M.~G.
\newblock Bootstrap latent-predictive representations for multitask
  reinforcement learning.
\newblock \emph{arXiv preprint arXiv:2004.14646}, 2020.

\bibitem[Guo et~al.(2018)Guo, Azar, Piot, Pires, and Munos]{guo2018neural}
Guo, Z.~D., Azar, M.~G., Piot, B., Pires, B.~A., and Munos, R.
\newblock Neural predictive belief representations.
\newblock \emph{arXiv preprint arXiv:1811.06407}, 2018.

\bibitem[Gutmann \& Hyv{\"a}rinen(2010)Gutmann and Hyv{\"a}rinen]{infonce}
Gutmann, M. and Hyv{\"a}rinen, A.
\newblock Noise-contrastive estimation: A new estimation principle for
  unnormalized statistical models.
\newblock In \emph{International Conference on Artificial Intelligence and
  Statistics}, 2010.

\bibitem[Ha \& Schmidhuber(2018)Ha and Schmidhuber]{ha2018world}
Ha, D. and Schmidhuber, J.
\newblock World models.
\newblock \emph{arXiv preprint arXiv:1803.10122}, 2018.

\bibitem[Haarnoja et~al.(2018)Haarnoja, Zhou, Abbeel, and
  Levine]{haarnoja2018soft}
Haarnoja, T., Zhou, A., Abbeel, P., and Levine, S.
\newblock Soft actor-critic: Off-policy maximum entropy deep reinforcement
  learning with a stochastic actor.
\newblock In \emph{International Conference on Machine Learning}, 2018.

\bibitem[Hafner et~al.(2019)Hafner, Lillicrap, Fischer, Villegas, Ha, Lee, and
  Davidson]{hafner2018learning}
Hafner, D., Lillicrap, T., Fischer, I., Villegas, R., Ha, D., Lee, H., and
  Davidson, J.
\newblock Learning latent dynamics for planning from pixels.
\newblock In \emph{International Conference on Machine Learning}, 2019.

\bibitem[Hafner et~al.(2020)Hafner, Lillicrap, Ba, and
  Norouzi]{hafner2019dream}
Hafner, D., Lillicrap, T., Ba, J., and Norouzi, M.
\newblock Dream to control: Learning behaviors by latent imagination.
\newblock In \emph{International Conference on Learning Representations}, 2020.

\bibitem[He et~al.(2019)He, Fan, Wu, Xie, and Girshick]{kaiming2019moco}
He, K., Fan, H., Wu, Y., Xie, S., and Girshick, R.
\newblock Momentum contrast for unsupervised visual representation learning.
\newblock \emph{arXiv preprint arXiv:1911.05722}, 2019.

\bibitem[He et~al.(2020)He, Fan, Wu, Xie, and Girshick]{he2019momentum}
He, K., Fan, H., Wu, Y., Xie, S., and Girshick, R.
\newblock Momentum contrast for unsupervised visual representation learning.
\newblock In \emph{IEEE/CVF Conference on Computer Vision and Pattern
  Recognition}, 2020.

\bibitem[H{\'e}naff et~al.(2019)H{\'e}naff, Srinivas, De~Fauw, Razavi, Doersch,
  Eslami, and Oord]{henaff2019data}
H{\'e}naff, O.~J., Srinivas, A., De~Fauw, J., Razavi, A., Doersch, C., Eslami,
  S., and Oord, A. v.~d.
\newblock Data-efficient image recognition with contrastive predictive coding.
\newblock \emph{arXiv preprint arXiv:1905.09272}, 2019.

\bibitem[Hessel et~al.(2018)Hessel, Modayil, van Hasselt, Schaul, Ostrovski,
  Dabney, Horgan, Piot, Azar, and Silver]{hessel2017rainbow}
Hessel, M., Modayil, J., van Hasselt, H., Schaul, T., Ostrovski, G., Dabney,
  W., Horgan, D., Piot, B., Azar, M., and Silver, D.
\newblock Rainbow: Combining improvements in deep reinforcement learning.
\newblock In \emph{AAAI Conference on Artificial Intelligence}, 2018.

\bibitem[Hessel et~al.(2019)Hessel, Soyer, Espeholt, Czarnecki, Schmitt, and
  van Hasselt]{hessel2019multi}
Hessel, M., Soyer, H., Espeholt, L., Czarnecki, W., Schmitt, S., and van
  Hasselt, H.
\newblock Multi-task deep reinforcement learning with popart.
\newblock In \emph{AAAI Conference on Artificial Intelligence}, 2019.

\bibitem[Jaderberg et~al.(2017)Jaderberg, Mnih, Czarnecki, Schaul, Leibo,
  Silver, and Kavukcuoglu]{jaderberg2016reinforcement}
Jaderberg, M., Mnih, V., Czarnecki, W.~M., Schaul, T., Leibo, J.~Z., Silver,
  D., and Kavukcuoglu, K.
\newblock Reinforcement learning with unsupervised auxiliary tasks.
\newblock In \emph{International Conference on Learning Representations}, 2017.

\bibitem[Jaderberg et~al.(2019)Jaderberg, Czarnecki, Dunning, Marris, Lever,
  Castaneda, Beattie, Rabinowitz, Morcos, Ruderman, et~al.]{jaderberg2019human}
Jaderberg, M., Czarnecki, W.~M., Dunning, I., Marris, L., Lever, G., Castaneda,
  A.~G., Beattie, C., Rabinowitz, N.~C., Morcos, A.~S., Ruderman, A., et~al.
\newblock Human-level performance in 3d multiplayer games with population-based
  reinforcement learning.
\newblock \emph{Science}, 364\penalty0 (6443):\penalty0 859--865, 2019.

\bibitem[Kalashnikov et~al.(2018)Kalashnikov, Irpan, Pastor, Ibarz, Herzog,
  Jang, Quillen, Holly, Kalakrishnan, Vanhoucke, et~al.]{kalashnikov2018qt}
Kalashnikov, D., Irpan, A., Pastor, P., Ibarz, J., Herzog, A., Jang, E.,
  Quillen, D., Holly, E., Kalakrishnan, M., Vanhoucke, V., et~al.
\newblock Qt-opt: Scalable deep reinforcement learning for vision-based robotic
  manipulation.
\newblock \emph{arXiv preprint arXiv:1806.10293}, 2018.

\bibitem[Kingma \& Welling(2013)Kingma and Welling]{kingma2013auto}
Kingma, D.~P. and Welling, M.
\newblock Auto-encoding variational bayes.
\newblock \emph{arXiv preprint arXiv:1312.6114}, 2013.

\bibitem[Kipf et~al.(2019)Kipf, van~der Pol, and Welling]{kipf2019contrastive}
Kipf, T., van~der Pol, E., and Welling, M.
\newblock Contrastive learning of structured world models.
\newblock \emph{arXiv preprint arXiv:1911.12247}, 2019.

\bibitem[Kostrikov et~al.(2020)Kostrikov, Yarats, and
  Fergus]{kostrikov2020image}
Kostrikov, I., Yarats, D., and Fergus, R.
\newblock Image augmentation is all you need: Regularizing deep reinforcement
  learning from pixels.
\newblock \emph{arXiv preprint arXiv:2004.13649}, 2020.

\bibitem[Laskin et~al.(2020{\natexlab{a}})Laskin, Lee, Stooke, Pinto, Abbeel,
  and Srinivas]{laskin_lee2020rad}
Laskin, M., Lee, K., Stooke, A., Pinto, L., Abbeel, P., and Srinivas, A.
\newblock Reinforcement learning with augmented data.
\newblock \emph{arXiv preprint arXiv:2004.14990}, 2020{\natexlab{a}}.

\bibitem[Laskin et~al.(2020{\natexlab{b}})Laskin, Srinivas, and
  Abbeel]{laskin2020curl}
Laskin, M., Srinivas, A., and Abbeel, P.
\newblock Curl: Contrastive unsupervised representations for reinforcement
  learning.
\newblock In \emph{International Conference on Machine Learning},
  2020{\natexlab{b}}.

\bibitem[Lee et~al.(2019)Lee, Nagabandi, Abbeel, and Levine]{lee2019stochastic}
Lee, A.~X., Nagabandi, A., Abbeel, P., and Levine, S.
\newblock Stochastic latent actor-critic: Deep reinforcement learning with a
  latent variable model.
\newblock \emph{arXiv preprint arXiv:1907.00953}, 2019.

\bibitem[Lee et~al.(2020)Lee, Fischer, Liu, Guo, Lee, Canny, and
  Guadarrama]{lee2020predictive}
Lee, K.-H., Fischer, I., Liu, A., Guo, Y., Lee, H., Canny, J., and Guadarrama,
  S.
\newblock Predictive information accelerates learning in rl.
\newblock \emph{Advances in Neural Information Processing Systems}, 33, 2020.

\bibitem[Levine et~al.(2016)Levine, Finn, Darrell, and Abbeel]{levine2015end}
Levine, S., Finn, C., Darrell, T., and Abbeel, P.
\newblock End-to-end training of deep visuomotor policies.
\newblock \emph{The Journal of Machine Learning Research}, 17\penalty0
  (1):\penalty0 1334--1373, 2016.

\bibitem[Mazoure et~al.(2020)Mazoure, Combes, Doan, Bachman, and
  Hjelm]{mazoure2020deep}
Mazoure, B., Combes, R. T.~d., Doan, T., Bachman, P., and Hjelm, R.~D.
\newblock Deep reinforcement and infomax learning.
\newblock \emph{arXiv preprint arXiv:2006.07217}, 2020.

\bibitem[Mnih et~al.(2015)Mnih, Kavukcuoglu, Silver, Rusu, Veness, Bellemare,
  Graves, Riedmiller, Fidjeland, Ostrovski, et~al.]{mnih2015human}
Mnih, V., Kavukcuoglu, K., Silver, D., Rusu, A.~A., Veness, J., Bellemare,
  M.~G., Graves, A., Riedmiller, M., Fidjeland, A.~K., Ostrovski, G., et~al.
\newblock Human-level control through deep reinforcement learning.
\newblock \emph{Nature}, 518\penalty0 (7540):\penalty0 529--533, 2015.

\bibitem[Mnih et~al.(2016)Mnih, Badia, Mirza, Graves, Lillicrap, Harley,
  Silver, and Kavukcuoglu]{a3c}
Mnih, V., Badia, A.~P., Mirza, M., Graves, A., Lillicrap, T., Harley, T.,
  Silver, D., and Kavukcuoglu, K.
\newblock Asynchronous methods for deep reinforcement learning.
\newblock In \emph{International Conference on Machine Learning}, 2016.

\bibitem[Schulman et~al.(2017)Schulman, Wolski, Dhariwal, Radford, and
  Klimov]{schulman2017proximal}
Schulman, J., Wolski, F., Dhariwal, P., Radford, A., and Klimov, O.
\newblock Proximal policy optimization algorithms.
\newblock \emph{arXiv preprint arXiv:1707.06347}, 2017.

\bibitem[Schwarzer et~al.(2020)Schwarzer, Anand, Goel, Hjelm, Courville, and
  Bachman]{schwarzer2020}
Schwarzer, M., Anand, A., Goel, R., Hjelm, R.~D., Courville, A., and Bachman,
  P.
\newblock Data-efficient reinforcement learning with momentum predictive
  representations.
\newblock \emph{arXiv preprint arXiv:2007.05929}, 2020.

\bibitem[Tassa et~al.(2018)Tassa, Doron, Muldal, Erez, Li, Casas, Budden,
  Abdolmaleki, Merel, Lefrancq, et~al.]{tassa2018deepmind}
Tassa, Y., Doron, Y., Muldal, A., Erez, T., Li, Y., Casas, D. d.~L., Budden,
  D., Abdolmaleki, A., Merel, J., Lefrancq, A., et~al.
\newblock Deepmind control suite.
\newblock \emph{arXiv preprint arXiv:1801.00690}, 2018.

\bibitem[van~den Oord et~al.(2018)van~den Oord, Li, and
  Vinyals]{oord2018representation}
van~den Oord, A., Li, Y., and Vinyals, O.
\newblock Representation learning with contrastive predictive coding.
\newblock \emph{arXiv preprint arXiv:1807.03748}, 2018.

\bibitem[Wayne et~al.(2018)Wayne, Hung, Amos, Mirza, Ahuja, Grabska-Barwinska,
  Rae, Mirowski, Leibo, Santoro, et~al.]{wayne2018unsupervised}
Wayne, G., Hung, C.-C., Amos, D., Mirza, M., Ahuja, A., Grabska-Barwinska, A.,
  Rae, J., Mirowski, P., Leibo, J.~Z., Santoro, A., et~al.
\newblock Unsupervised predictive memory in a goal-directed agent.
\newblock \emph{arXiv preprint arXiv:1803.10760}, 2018.

\bibitem[Yan et~al.(2020)Yan, Vangipuram, Abbeel, and Pinto]{yan2020learning}
Yan, W., Vangipuram, A., Abbeel, P., and Pinto, L.
\newblock Learning predictive representations for deformable objects using
  contrastive estimation.
\newblock \emph{arXiv preprint arXiv:2003.05436}, 2020.

\bibitem[Zagoruyko \& Komodakis(2016)Zagoruyko and
  Komodakis]{zagoruyko2016paying}
Zagoruyko, S. and Komodakis, N.
\newblock Paying more attention to attention: Improving the performance of
  convolutional neural networks via attention transfer.
\newblock \emph{arXiv preprint arXiv:1612.03928}, 2016.

\end{thebibliography}
\bibliographystyle{icml2021}

\clearpage
\appendix
\onecolumn
\section{Appendix}

\subsection{Algorithms}

\begin{algorithm}
\caption{\newline Online RL with decoupled ATC encoder (\textcolor{blue}{steps distinct from end-to-end RL in blue})}
\label{algo:decoupled_rl}
\begin{algorithmic}[1]
\Require $\theta_{ATC}, \phi_\pi$
\Comment ATC model parameters (encoder $f_\theta$ thru contrast $W$), policy parameters

\State $\color{blue} \mathcal{S} \gets \{\}$  \Comment{replay buffer of observations}
\State $\color{blue} \bar{\theta}_{ATC} \gets \theta_{ATC}$
\Comment initialize momentum encoder (conv and linear only)
\Repeat

\State Sample environment and policy, through encoder:
\For{1 to m}
\Comment{a minibatch}
\State $a\sim\pi(\cdot|f_\theta(s);\phi), s'\sim T(s,a), r\sim R(s,a,s')$
\State $\color{blue} \mathcal{S}\gets\mathcal{S}\cup \{s\}$
\Comment store observations (delete oldest if full)
\State $s \gets s'$
\EndFor

\State Update policy by given RL formula:
\Comment on- or off-policy
\For{1 to n}
\Comment given number RL updates per minibatch
\State $\quad \phi_\pi \gets \phi_\pi + RL(s,a,s',r;\phi_\pi)$
\Comment stop gradient into encoder
\EndFor

\State Update encoder (and contrastive model) by ATC:
\For{1 to p}
\State $\color{blue} s,s_+\sim \mathcal{S}$ 
\Comment sample observations: anchors and positives
\State $\color{blue} \theta_{ATC} \gets \theta_{ATC} - \lambda_{ATC} \nabla_{\theta_{ATC}}\mathcal{L}^{ATC}(s,s_+)$
\Comment ATC gradient update
\State $\color{blue} \bar{\theta}_{ATC} \gets (1-\tau)\bar{\theta}_{ATC} + \tau \theta_{ATC}$
\Comment update momentum encoder (conv and linear only)
\EndFor

\Until{converged}
\State \textbf{return} Encoder $f_\theta$ and policy $\pi_\phi$
\end{algorithmic}
\end{algorithm}

\clearpage
\subsection{Additional Figures}

\begin{figure}[h]
    \centering
    \includegraphics[width=0.95\textwidth]{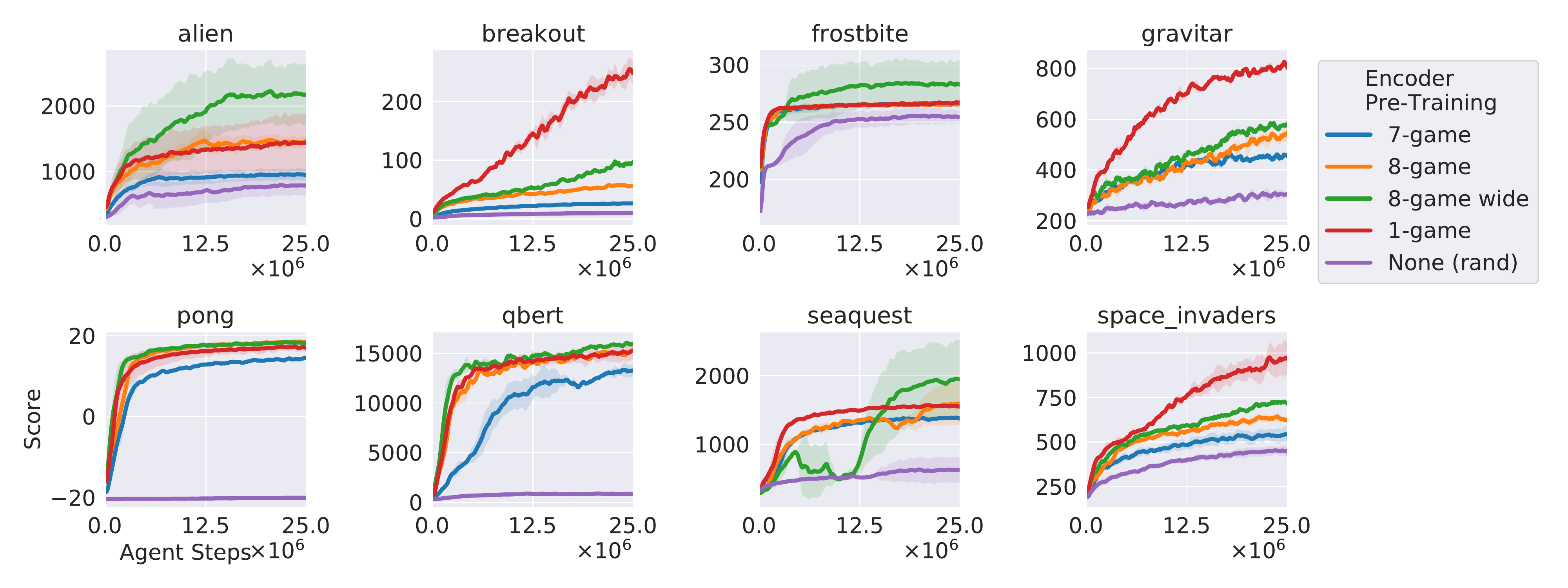}
    \caption{\small RL using multi-task encoders (all with weights frozen) for eight Atari games gives mixed performance, partially improved by increased network capacity (8-game-wide).  Training on 7 games and testing on the held-out one yields diminished but non-zero performance, showing some limited feature transfer between games.}
    \label{fig:atari_ppo_multi}
\end{figure}

\begin{figure}[h]
    \centering
    \includegraphics[width=0.9\textwidth]{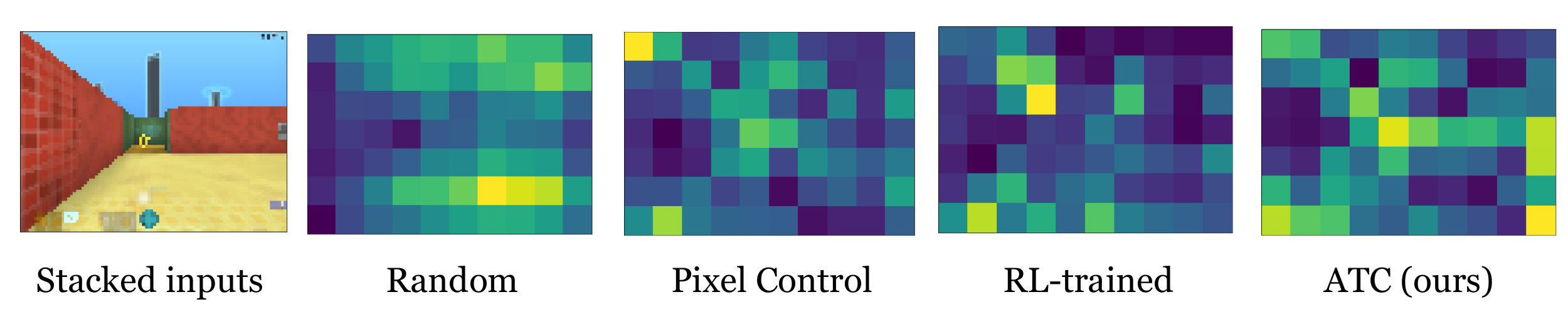}
    \caption{Attention map in \textsc{LASERTAG}. UL encoder with pixel control focuses on the score, while UL encoder with the proposed ATC focuses properly on the coin similar to RL-trained encoder.}
    \label{fig:lasertag_attention_2}
\end{figure}

\begin{figure}[H]
    \centering
    \includegraphics[width=0.9\textwidth]{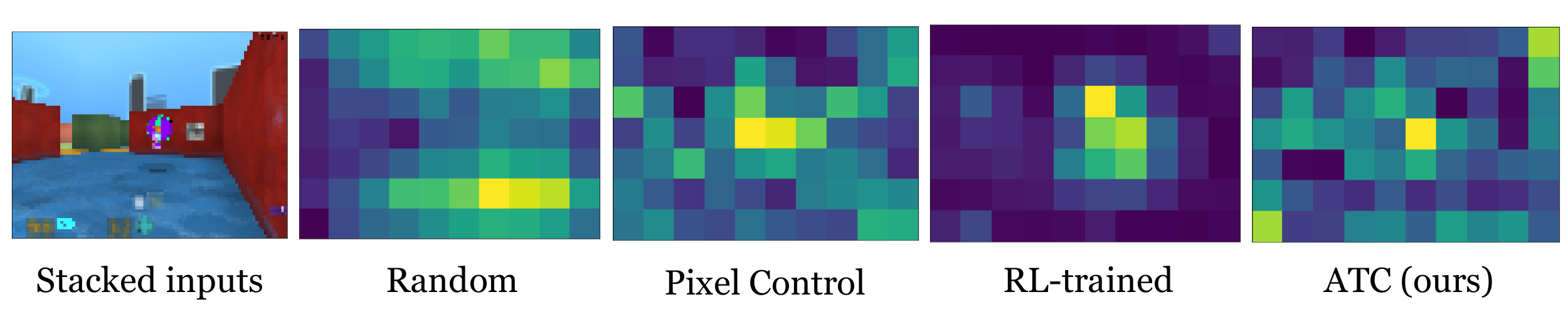}
    \caption{Attention map in the \textsc{LASERTAG} which shows that UL encoders focus properly on the enemy similar to RL-trained encoder.}
    \label{fig:lasertag_attention_1}
\end{figure}

\clearpage
\subsection{RL Settings}

\begin{table}[h]
\caption{DMControl, RAD-SAC Hyperparameters.}
\label{table:dmcontrol_sac_params}
\vskip 0.15in
\begin{center}
\begin{small}
\begin{sc}
\begin{tabular}{rl}
\toprule
Hyperparameter & Value \\
\midrule
Observation rendering    & $(84,84)$, RGB  \\ 
Random Shift Pad    & $\pm 4$  \\ 
Replay buffer size    & \num{1e5} \\ 
Initial steps    & \num{1e4}  \\ 
Stacked frames    & $3$  \\ 
Action repeat    & $2$ (Finger, Walker) \\
 & $8$ (Cartpole)\\
 & $4$ (rest) \\
Optimizer    & Adam  \\ 
$(\beta_1,\beta_2) \rightarrow (f_\theta, \pi_\psi, Q_\phi)$   & $(.9,.999)$  \\
$(\beta_1,\beta_2) \rightarrow (\alpha)$   & $(.5,.999)$  \\
Learning rate $(f_\theta, \pi_\psi, Q_\phi)$     & \num{2e-4} (Cheetah) \\
 & \num{1e-3} (rest) \\ 
Learning rate ($\alpha$) & \num{1e-4} \\
Batch Size    & $512$ (Cheetah, Pendulum) \\
 & 256 (rest)  \\ 
$Q$ function EMA $\tau$ & $0.01$ \\
Critic target update freq & $2$ \\
Convolution filters & $[32, 32, 32, 32]$ \\
Convolution strides & $[2, 2, 2, 1]$ \\
Convolution filter size & $3$ \\
Encoder EMA $\tau$ & $0.05$ \\
Latent dimension & $50$ \\
Hidden units (MLP)    & $[1024, 1024]$  \\
Discount $\gamma$ & $.99$ \\
Initial temperature & $0.1$ \\
\bottomrule
\end{tabular}
\end{sc}
\end{small}
\end{center}
\vskip -0.1in
\end{table}

\begin{table}[h]
\caption{Atari, PPO Hyperparameters.}
\label{table:atari_ppo_params}
\vskip 0.15in
\begin{center}
\begin{small}
\begin{sc}
\begin{tabular}{rl}
\toprule
Hyperparameter & Value \\
\midrule
Observation rendering    & $(84,84)$, Grey  \\ 
Stacked frames    & $4$  \\ 
Action repeat    & $4$ \\
Optimizer    & Adam  \\ 
Learning rate & \num{2.5e-4}  \\
Parallel Environments & 16 \\
Sampling interval & 128 \\
Likelihood ratio clip, $\eps$ & 0.1 \\
PPO Epochs & 4 \\
PPO Minibatches & 4 \\
Convolution filters & $[32, 64, 64]$ \\
Convolution filter sizes & $[8, 4, 3]$ \\
Convolution strides & $[4, 2, 1]$ \\
Hidden units (MLP) & $[512]$  \\
Discount $\gamma$ & $.99$ \\
Generalized Advantage Estimation $\lambda$ & $0.95$ \\
Learning rate annealing & linear \\
Entropy bonus coefficient & 0.01 \\
Episodic lives & False \\
Repeat action probability & $0.25$ \\
Reward clipping & $\pm1$ \\
Value loss coefficient & $1.0$ \\
\bottomrule
\end{tabular}
\end{sc}
\end{small}
\end{center}
\vskip -0.1in
\end{table}

\begin{table}[h]
\caption{DMLab, PPO Hyperparameters.}
\label{table:dmlab_ppo_params}
\vskip 0.15in
\begin{center}
\begin{small}
\begin{sc}
\begin{tabular}{rl}
\toprule
Hyperparameter & Value \\
\midrule
Observation rendering    & $(72,96)$, RGB  \\ 
Stacked frames    & $1$  \\ 
Action repeat    & $4$ \\
Optimizer    & Adam  \\ 
Learning rate & \num{2.5e-4}  \\
Parallel Environments & 16 \\
Sampling interval & 128 \\
Likelihood ratio clip, $\eps$ & 0.1 \\
PPO Epochs & 1 \\
PPO Minibatches & 2 \\
Convolution filters & $[32, 64, 64, 64]$ \\
Convolution filter sizes & $[8, 4, 3, 3]$ \\
Convolution strides & $[4, 2, 1, 1]$ \\
Hidden units (LSTM) & $[256]$  \\
Skip connections & conv 3, 4; LSTM \\
Discount $\gamma$ & $.99$ \\
Generalized Advantage Estimation $\lambda$ & $0.97$ \\
Learning rate annealing & none \\
Entropy bonus coefficient & $0.01$ (Explore) \\
 & $0.0003$ (Lasertag) \\
Value loss coefficient & $0.5$ \\
\bottomrule
\end{tabular}
\end{sc}
\end{small}
\end{center}
\vskip -0.1in
\end{table}

\clearpage
\subsection{Online ATC Settings}

\begin{table}[h]
\caption{Common ATC Hyperparameters.}
\label{table:dmcontrol_params}
\vskip 0.15in
\begin{center}
\begin{small}
\begin{sc}
\begin{tabular}{rl}
\toprule
Hyperparameter & Value \\
\midrule
Random shift pad & $\pm4$ \\
Learning rate & \num{1e-3} \\
Learning rate annealing & cosine \\
Target update interval & $1$ \\
Target update $\tau$ & $0.01$ \\
Predictor hidden sizes, $h_\psi$ & $[512]$ \\
Replay buffer size & \num{1e5} \\
\bottomrule
\end{tabular}
\end{sc}
\end{small}
\end{center}
\vskip -0.1in
\end{table}

\begin{table}[h]
\caption{DMControl ATC Hyperparameters.}
\label{table:dmcontrol_atc_params}
\vskip 0.15in
\begin{center}
\begin{small}
\begin{sc}
\begin{tabular}{rl}
\toprule
Hyperparameter & Value \\
\midrule
Random shift probability & $1$ \\
Batch size & as RL (individual observations)\\
Temporal shift, $k$ & 1 \\
Min agent steps to UL & \num{1e4} \\
Min agent steps to RL & \num{1e4} \\
UL update schedule & as RL \\
 & ($2$x Cheetah) \\
Latent size & $128$ \\
\bottomrule
\end{tabular}
\end{sc}
\end{small}
\end{center}
\vskip -0.1in
\end{table}

\begin{table}[h]
\caption{Atari ATC Hyperparameters.}
\label{table:atari_params}
\vskip 0.15in
\begin{center}
\begin{small}
\begin{sc}
\begin{tabular}{rl}
\toprule
Hyperparameter & Value \\
\midrule
Random shift probability & $0.1$ \\
Batch size & 512 (32 trajectories of 16 time steps) \\
Temporal shift, $k$ & 3 \\
Min agent steps to UL & \num{5e4} \\
Min agent steps to RL & \num{1e5} \\
UL update schedule & Annealed quadratically from 6 per sampler iteration \\
 & (\num{1e4} once at \num{1e5} steps for weight initialization) \\
Latent size & $256$ \\
\bottomrule
\end{tabular}
\end{sc}
\end{small}
\end{center}
\vskip -0.1in
\end{table}

\begin{table}[!htb]
\caption{DMLab ATC Hyperparameters.}
\label{table:dmlab_params}
\vskip 0.15in
\begin{center}
\begin{small}
\begin{sc}
\begin{tabular}{rl}
\toprule
Hyperparameter & Value \\
\midrule
Random shift probability & $1$ \\
Batch size & 512 (individual observations) \\
Temporal shift, $k$ & 3 \\
Min agent steps to UL & \num{5e4} \\
Min agent steps to RL & \num{1e5} \\
UL update schedule & 2 per sampler iteration \\
Latent size & $256$ \\
\bottomrule
\end{tabular}
\end{sc}
\end{small}
\end{center}
\vskip -0.1in
\end{table}

\clearpage
\subsection{Offline Pre-Training Details}
We conducted coarse hyperparameter sweeps to tune each competing UL algorithm.  In all cases, the best setting is the one shown in our comparisons.  

When our VAEs include a time difference between input and reconstruction observations, we include one hidden layer with action additionally input between the encoder and decoder.  We tried both 1.0 and 0.1 KL-divergence weight in the VAE loss, and found 0.1 to perform better in both DMControl and Atari.

\paragraph{DMControl}  For the VAE, we experimented with 0 and 1 time step difference between input and reconstruction target observations and training for either \num{1e4} or \num{5e4} updates.  The best settings were 1-step temporal, and \num{5e4} updates, with batch size 128.  ATC used 1-step temporal, \num{5e4} updates (although this can be significantly decreased), and batch size 256 (including \textsc{Cheetah}).  The pretraining data set consisted of the first \num{5e4} transitions from a RAD-SAC agent learning each task, including \num{5e3} random actions.  Within this span, \textsc{Cartpole} and \textsc{Ball\_in\_Cup} learned completely, but \textsc{Walker} and \textsc{Cheetah} reached average returns of 514 and 630, respectively (collected without the compressive convolution).

\paragraph{DMLab}  For Pixel Control, we used the settings from \cite{hessel2019multi} (see the appendix therein), except we used only empirical returns, computed offline (without bootstrapping).  For CPC, we tried training batch shapes, $batch\times time$ in (64, 8), (32, 16), (16, 32), and found the setting with rollouts of length 16 to be best.  We contrasted all elements of the batch against each other, rather than only forward constrasts.  In all cases we also used 16 steps to warmup the LSTM.   For all algorithms we tried learning rates \num{3e-4} and \num{1e-3} and both \num{5e4} and \num{1.5e5} updates.  For ATC and CPC, the lower learning rate and higher number of updates helped in \textsc{Lasertag} especially.  The pretraining data was \num{125e3} samples from partially trained RL agents receiving average returns of 127 and 6 in \textsc{Explore\_Goal\_Locations\_Small} and \textsc{Lasertag\_Three\_Opponents\_Small}, respectively.

\paragraph{Atari} For the VAE, we experimented with 0, 1, and 3 time step difference between input and reconstruction target, and found 3 to work best.  For ST-DIM we experimented with 1, 3, and 4 time steps differences, and batch sizes from 64 to 256, learning rates \num{1e-3} and \num{5e-4}.  Likewise, 3-step delay worked best.  For the inverse model, we tried 1- and 3-step predictions, with 1-step working better overall, and found random shift augmentation to help.  For pixel control, we used the settings in \cite{jaderberg2016reinforcement}, again with full empirical returns.  We ran each algorithm for up to \num{1e5} updates, although final ATC results used \num{5e4} updates.  We ran each RL agent with and without observation normalization on the latent image and observed no difference in performance.  Pretraining data was \num{125e3} samples sourced from the replay buffer of DQN agents trained for \num{15e6} steps with epsilon-greedy $\epsilon=0.1$.  Evaluation scores were:

\begin{table}[h]
\caption{Atari Pre-Training Data Source Agents.}
\label{table:atari_pretrain_data}
\vskip 0.15in
\begin{center}
\begin{small}
\begin{sc}
\begin{tabular}{rl}
\toprule
Game & Evaluation Score \\
\midrule
Alien & $1,800$ \\
Breakout & $279$ \\
Frostbite & $1,400$ \\
Gravitar & $390$ \\
Pong & $18$ \\
QBert & $8,800$ \\
Seaquest & $11,000$ \\
Space Invaders & $1,200$ \\
\bottomrule
\end{tabular}
\end{sc}
\end{small}
\end{center}
\vskip -0.1in
\end{table}

\clearpage

\end{document}